\documentclass[11pt,a4paper]{article}
\usepackage[hyperref]{acl2017}
\usepackage{times}
\usepackage{latexsym}
\usepackage{graphicx}
\usepackage{hhline}
\usepackage{booktabs}
\usepackage{multirow}
\usepackage{pgfplots}
\pgfplotsset{compat=1.9}
\usepackage{color}
\usepackage{float}
\usepackage{caption}
\usepackage{amssymb}
\usepackage{amsmath}
\usepackage{breqn}
\usepackage{verbatim}
\usepackage{svg}
\usepackage{pifont}
\usepackage{textcomp}
\usepackage{lipsum}
\usepackage{microtype}

\newcommand\blfootnote[1]{%
  \begingroup
  \renewcommand\thefootnote{}\footnote{#1}%
  \addtocounter{footnote}{-1}%
  \endgroup
}

\usepackage{url}

\aclfinalcopy % Uncomment this line for the final submission
\title{Combating Human Trafficking with Deep Multimodal Models}

\author{%
  Edmund Tong* \\
  Language Technologies Institute \\
  Carnegie Mellon University \\
  {\tt edtong@cmu.edu} \\\And
  Amir Zadeh* \\
  Language Technologies Institute \\
  Carnegie Mellon University \\
  {\tt abagherz@cs.cmu.edu} \\\AND
  Cara Jones \\
  Marinus Analytics, LLC \\\\
  {\tt cara@marinusanalytics.com } \\\And
  Louis-Philippe Morency \\
  Language Technologies Institute \\
  Carnegie Mellon University \\
  {\tt morency@cs.cmu.edu} \\
}

\date{}

\begin{document}

\maketitle
\blfootnote{\hspace{-0.8em}*\,Authors contributed equally.}

\par\noindent
\begin{abstract}
  Human trafficking is a global epidemic affecting millions of people across the planet.
  Sex trafficking, the dominant form of human trafficking, has seen a significant rise mostly due to the abundance of escort websites, where human traffickers can openly advertise among at-will escort advertisements.
  In this paper, we take a major step in the automatic detection of advertisements suspected to pertain to human trafficking.
  We present a novel dataset called Trafficking-10k, with more than 10,000~advertisements annotated for this task.
  The dataset contains two sources of information per advertisement: text and images.
  For the accurate detection of trafficking advertisements, we designed and trained a deep multimodal model called the Human Trafficking Deep Network (HTDN). 
\end{abstract}
\section{Introduction}
Human trafficking ``a crime that shames us all'' \cite{unodc2008}, has seen a steep rise in the United States since~2012.
The number of cases reported rose from 3,279 in~2012 to 7,572 in~2016---more than doubling over the course of five years \cite{hotline2017}. 
%Sex trafficking is the most common form of human trafficking and is a global epidemic affecting millions of men, women, and children each year \cite{mccarthy2014human}.
% The above source does not seem to support the claim. Indeed, it states the following:
% ``A focus on sex trafficking means that these other forms of exploitation often go unrecognized and other areas of policy that could lower the risks of exploitation tend to be ignored.''
Sex trafficking is a form of human trafficking, and is a global epidemic affecting millions of people each year \cite{mccarthy2014human}.
%The UN estimates that it is the world's fastest growing type of organized crime and the third most common in the world.%
%\footnote{\url{http://goo.gl/UaX0G7}}
% Removed: Source does not support claim.
Victims of sex trafficking are subjected to coercion, force, and control, and are not able to ask for help.
Put plainly, sex trafficking is modern-day slavery and is one of the top priorities of law enforcement agencies at all levels.

A major advertising ground for human traffickers is the World Wide Web.
The Internet has brought traffickers the ability to advertise online and has fostered the growth of numerous adult escort sites.
Each day, there are tens of thousands of Internet advertisements posted in the United States and Canada that market commercial sex.
Hiding among the noise of at-will adult escort ads are ads posted by sex traffickers.
Often long undetected, trafficking rings and escort websites form a profit cycle that fuels the increase of both trafficking rings and escort websites.
% Above claim unsubstantiated.

For law enforcement, this presents a significant challenge: how should we identify advertisements that are associated with sex trafficking?
Police have limited human and technical resources, and manually sifting through thousands of ads in the hopes of finding something suspicious is a poor use of those resources, even if they know what they are looking for.
Leveraging state-of-the-art machine learning approaches in~Natural Language Processing and computer vision to detect and report advertisements suspected of trafficking is the main focus of our work.
In other words, we strive to find the victims and perpetrators of trafficking who hide in plain sight in the massive amounts of data online.
By narrowing down the number of advertisements that law enforcement must sift through, we endeavor to provide a real opportunity for law enforcement to intervene in the lives of victims.
However, there are non-trivial challenges facing this line of research:

\textbf{Adversarial Environment.}
Human trafficking rings are aware that law enforcement monitors their online activity.
Over the years, law enforcement officers have populated lists of keywords that frequently occur in trafficking advertisements.
However, these simplistic queries fail when traffickers use complex obfuscation.
Traffickers, again aware of this, move to new keywords to blend in with the at-will escort advertisements.
This trend creates an adversarial environment for any machine learning system that attempts to find trafficking rings hiding in plain sight.

\textbf{Defective Language Compositionality.}
Online escort advertisements are difficult to analyze, because they lack grammatical structures such as constituency.
Therefore, any form of inference must rely more on context than on grammar.
This presents a significant challenge to the NLP community.
Furthermore, the majority of the ads contain emojis and non-English characters.

\textbf{Generalizable Language Context.}
Machine learning techniques can easily learn unreliable cues in training sets such as phone numbers, keywords, and other forms of semantically unreliable discriminators to reduce the training loss.
Due to limited similarity between the training and test data due to the large number of ads available online, relying on these cues is futile.
Learned discriminative features should be generalizable and model semantics of trafficking.

\textbf{Multimodal Nature.}
Escort advertisements are composed of both textual and visual information.
Our model should treat these features interdependently. 
For instance, if the text indicates that the escort is in a hotel room, our model should consider the effect that such knowledge may have on the importance of certain visual features.
% TODO: Consider \emph{}asizing both ``text'' and ``visual.''

We believe that studying human trafficking advertisements can be seen as a fundamental challenge to the NLP, computer vision, and machine learning communities dealing with language and vision problems.
In this paper, we present the following contributions to this research direction.
First, we study the language and vision modalities of the escort advertisements through deep neural modeling.
%While there are similarities in the text and images used in these advertisements with text and images used in tweets and reviews, there are fundamental differences that are distinguishable to research in this area.
Second, we take a significant step in automatic detection of advertisements suspected of sex trafficking.
While previous methods \cite{dubrawski2015leveraging} have used simplistic classifiers, we build an end-to-end-trained multimodal deep model called the Human Trafficking Deep Network (HTDN).
The HTDN uses information from both text and images to extract cues of human trafficking, and shows outstanding performance compared to previously used models.
Third, we present the first rigorously annotated dataset for detection of human trafficking, called Trafficking-10k, which includes more than 10,000 trafficking ads labeled with likelihoods of having been posted by traffickers.%
\footnote{%
  Due to the sensitive nature of this dataset, access can only be granted by emailing Cara Jones.
  Different levels of access are provided \emph{only} to scientific community.
}
%https://humantraffickinghotline.org/sites/default/files/NHTRC%202015%20United%20States%20Report%20-%20USA%20-%2001.01.15%20-%2012.31.15_OTIP_Edited_06-09-16.pdf
\section{Related Works}

Automatic detection of human trafficking has been a relatively unexplored area of machine learning research.
Very few machine learning approaches have been proposed to detect signs of human trafficking online.
Most of these approaches use simplistic methods such as multimedia matching \cite{zhou2016multimedia}, text-based filtering classifiers such as random forests, logistic regression, and SVMs \cite{dubrawski2015leveraging}, and named-entity recognition to isolate the instances of trafficking \cite{nagpal2015entity}.
Studies have suggested using statistical methods to find keywords and signs of trafficking from data to help law enforcement agencies \cite{kennedy2012predictive} as well as adult content filtering using textual information \cite{zhou2016multimedia}.

Multimodal approaches have gained popularity over the past few years.
These multimodal models have been used for medical purposes, such as detection of suicidal risk, PTSD and depression \cite{scherer2016self,venek2016adolescent,yu2013multimodal,valstar2016avec}; sentiment analysis \cite{zadeh2016multimodal,poria2016convolutional,zadeh2016mosi}; emotion recognition \cite{poria2017review}; image captioning and media description \cite{you2016image, donahue2015long}; question answering \cite{antol2015vqa}; and multimodal translation \cite{specia2016shared}.

To the best of our knowledge, this paper presents the first multimodal and deep model for detection of human trafficking. 
\section{Trafficking-10k Dataset}

In this section, we present the dataset for our studies.
We formalize the problem of recognizing sex trafficking as a machine learning task.
The input data is text and images; this is mapped to a measure of how suspicious the advertisement is with regards to human trafficking.%
\begin{figure}[t!]
	\centering
	\includegraphics[width=0.5\textwidth]{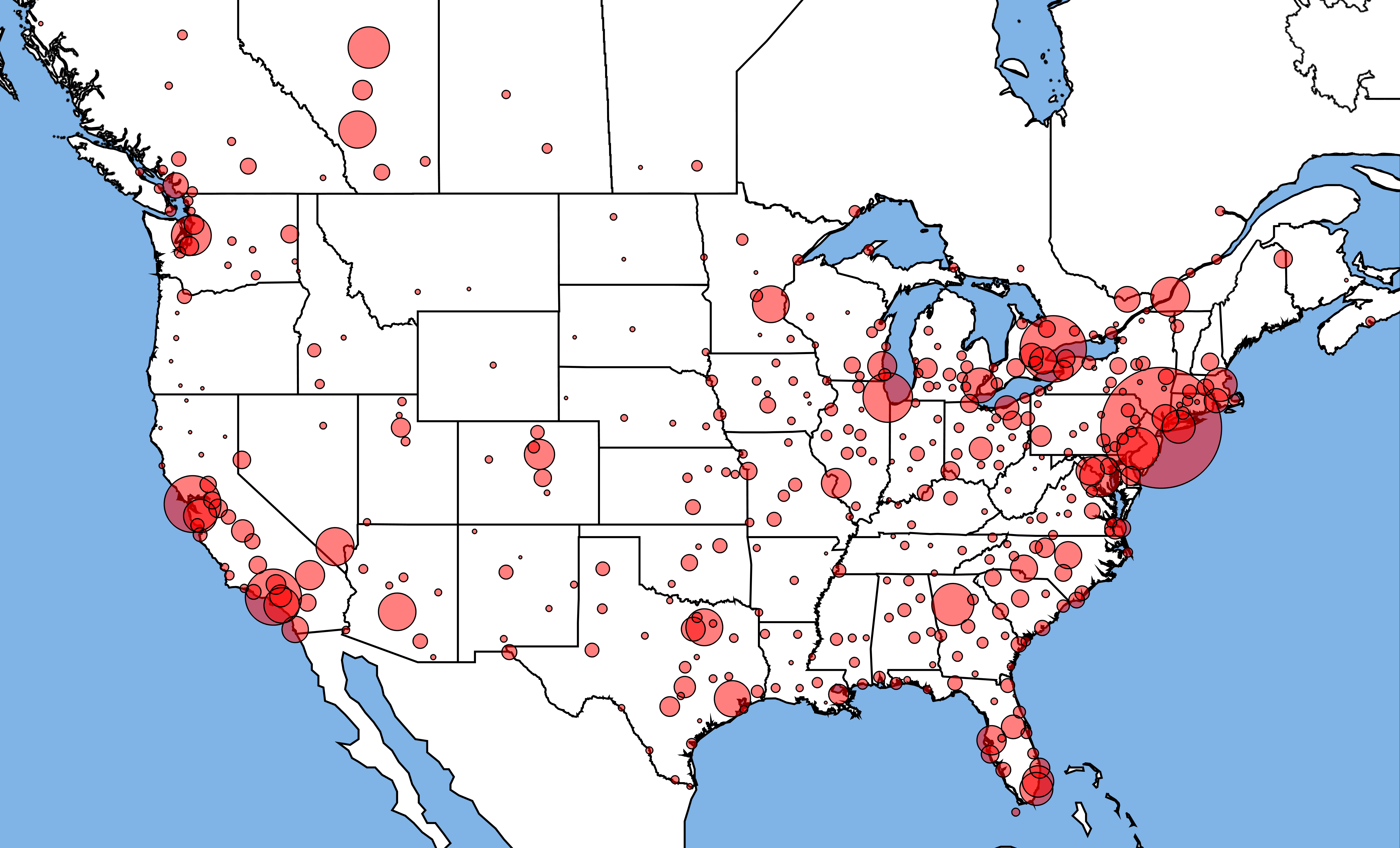}
	\caption{Distribution of advertisements in Trafficking-10k dataset across United States and Canada.}
	\label{fig:usmap}
\end{figure} %
\subsection{Data Acquisition and Preprocessing}
A subset of 10,000 ads were sampled randomly from a large cache of escort ads for annotation in Trafficking-10k dataset.
The distribution of advertisements across the United States and Canada is shown in Figure~\ref{fig:usmap}, which indicates the diversity of advertisements in Trafficking-10k. This diversity ensures that models trained on Trafficking-10k can be applicable nationwide.
%Furthermore, it enables studies that target differences in dynamics of trafficking depending on geographic location.
% Above sentence struck because T10k does not include geographic data.
The 10,000 collected ads each consist of text and zero or more images.
The text in the dataset is in plain text format, derived by stripping the HTML tags from the raw source of the ads.
%The text was derived from the raw HTML of advertisements;
%As the first step in preprocessing text, the HTML tags are stripped and only the content of the ads are considered important.
The set of characters in each advertisement is encoded as UTF-8, because there is ample usage of smilies and non-English characters.
%Unlike HTML tags, these characters are informative since they are semantically meaningful, such as dollar sign smiley representing only-cash exchange.
Advertisements are truncated to the first 184~words, as this covers more than 90\% of the ads. Images are resized to $224 \times 224$ pixels with RGB channels.

\subsection{Trafficking Annotation}
Detecting whether or not an advertisement is suspicious requires years of practice and experience in working closely with law enforcement.
As a result, annotation is a highly complicated and expensive process, which cannot be scaled using crowdsourcing.
In our dataset, annotation is carried out by two expert annotators, each with at least five years of experience, in detection of human trafficking and another annotator with one year of experience.
In our dataset, annotations were done by three experts.
One expert has over a year of experience, and the other two have over five years of experience in the human trafficking domain.
%To calculate the inter-annotator agreement, each annotator is given a set of 1,000 similar ads to annotate.
% TODO: This is wrong. Edmund to fix
To calculate the inter-annotator agreement, each annotator is given the same set of 1000 ads to annotate and the nominal agreement is found: there was a 83\% pairwise agreement (0.62 Krippendorff's alpha).
Also, to make sure that annotations are generalizable across the annotators and law enforcement officers, two law enforcement officers annotated, respectively, a subset of 500 and 100 of the advertisements. 
We found a 62\% average pairwise agreement (0.42 Krippendorff's alpha) with our annotators.
This gap is reasonable, as law enforcement officers only have experience with local advertisements, while Trafficking-10k annotators have experience with cases across the United States. %

Annotators used an annotation interface specifically designed for the Trafficking-10k dataset.
In the annotation interface, each advertisement was displayed on a separate webpage.
The order of the advertisements is determined uniformly randomly, and annotators were unable to move to the next advertisement without labeling the current one.
For each advertisement, the annotator was presented with the question: ``In your opinion, would you consider this advertisement suspicious of human trafficking?''
The annotator is presented with the following options: ``Certainly no,'' ``Likely no,'' ``Weakly no,'' ``Unsure,''\footnote{This option is greyed out for 10~seconds to encourage annotators to make an intuitive decision.} ``Weakly yes,'' ``Likely yes,'' and ``Certainly yes.''
Thus, the degree to which advertisements are suspicious is quantized into seven levels.

\subsection{Analysis of Language}

The language used in these advertisements introduces fundamental challenges to the field of NLP.
The nature of the textual content in these advertisements raises the question of how we can make inferences in a linguistic environment with a constantly evolving lexicon.
Language used in the Trafficking-10k dataset is highly inconsistent with standard grammar.
Often, words are obfuscated by emojis and symbols.
The word ordering is inconsistent, and there is rarely any form of constituency.
This form of language is completely different from spoken and written English.
These attributes make escort advertisements appear somewhat similar to tweets, specifically since these ads are normally short (more than~90\% of the ads have at most 184~words).
Another point of complexity in these advertisements is the high number of unigrams, due to usage of uncommon words and obfuscation.
On top of unigram complexity, advertisers continuously change their writing pattern, making this problem more complex.

\begin{figure}[t!]
  \centering
  \begin{tikzpicture}
    \begin{axis}
      [
        width=0.45\textwidth,
        height=2in,
        bar width=6.5pt,
        ybar stacked,
        legend pos=north west,
        title={Advertisement lengths},
        xlabel={Number of unigrams},
        xtick={0,40,...,200},
        extra x ticks={230},
        extra x tick labels={$+$},
      ]
      \addplot table[x=max, y=pos] {data/advertisement_lengths_bins.dat};
      \addplot table[x=max, y=neg] {data/advertisement_lengths_bins.dat};
      \addlegendentry{Positive}
      \addlegendentry{Negative}
    \end{axis}
  \end{tikzpicture}
  \caption{Distribution of the length of advertisements in Trafficking-10k. There is no significant difference between positive and negative cases purely based on length.}
  \label{fig:adlen}
\end{figure}
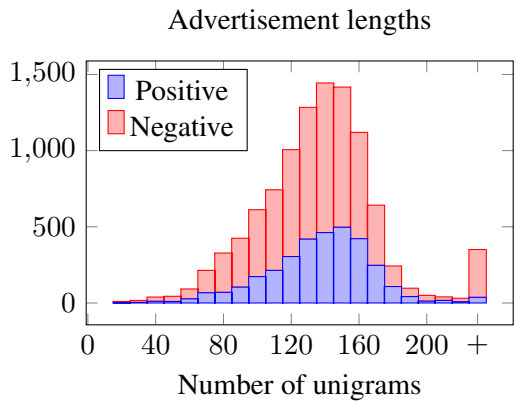
\subsection{Dataset Statistics}

There are 106,954 distinct unigrams, 353,324 distinct bigrams, and 565,403 trigrams in the Trafficking-10k dataset.
There are 60,337 images.
The total number of distinct characters including whitespace, punctuations, and hex characters is~182.
The average length of an ad is~137 words, with a standard deviation of~74, median~133.
The shortest advertisement has 7~unigrams, and the longest advertisement has 1810~unigrams.
There are of 106,954 distinct unigrams, 353,324 distinct bigrams and 565,403 trigrams in the Trafficking-10k dataset.
The average number of images in an advertisement is~5.9; the median is~5, the minimum is~0, and the maximum is~90.

The length of \emph{suspected} advertisements is 134 unigrams; the standard deviation is 39, the minimum is 12, and the maximum is 666.
The length of \emph{non-suspected} ads is 141; the standard deviation is 85, the minimum is 7, and the maximum is 1810.
The total number of suspected ads is~3257; and the total number of non-suspected ads is~6992.
Figure~\ref{fig:adlen} shows the histogram of number of ads based on their length. Both the positive and negative distributions are similar. This means that there is no obvious length difference between the two classes.
Most of the ads have a length of~80--180 words.

\begin{figure*}[t]
	\centering
	\includegraphics[width=0.9\textwidth]{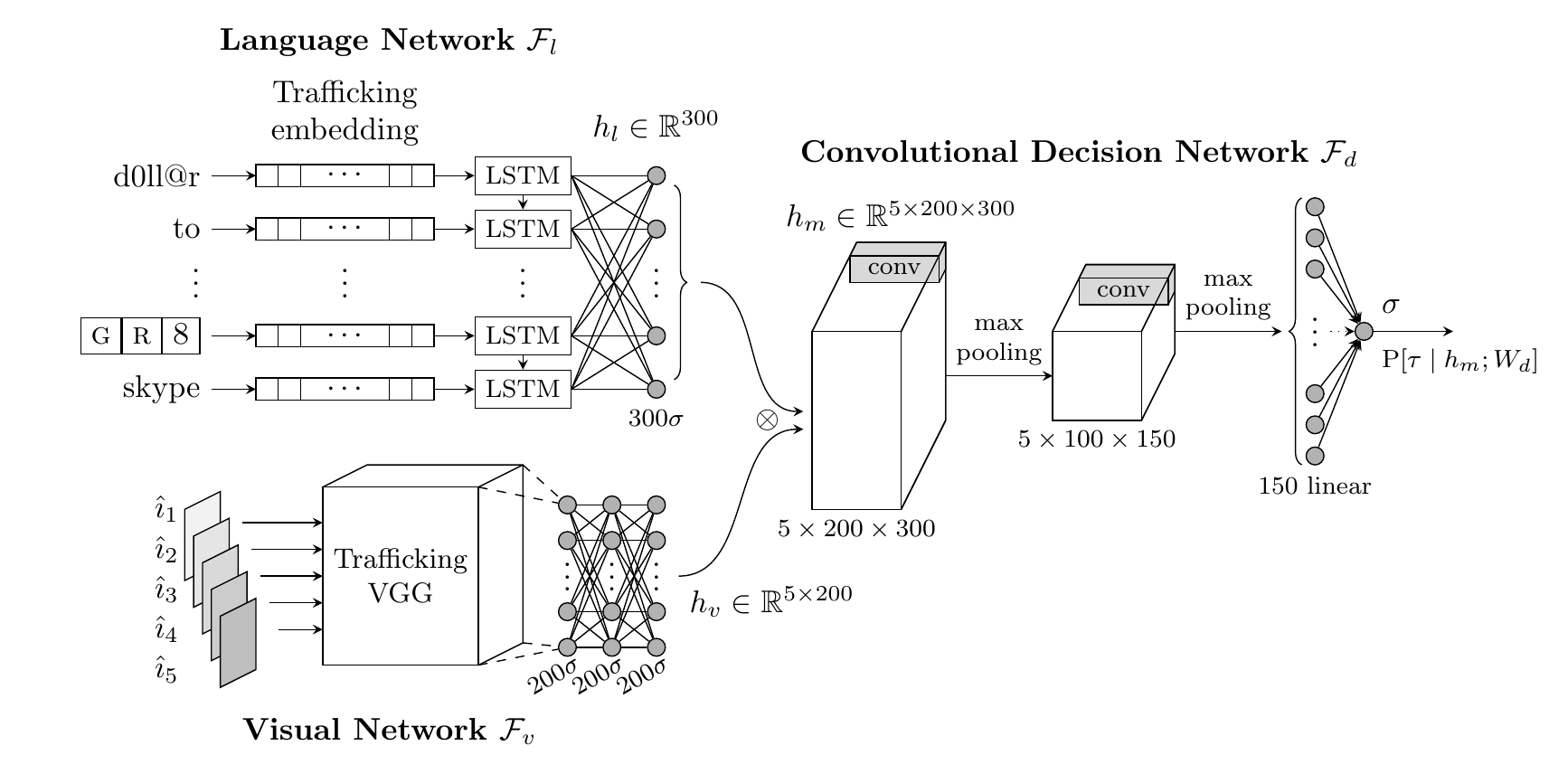}
	\caption{Overview of our proposed Human Trafficking Deep Network (HTDN). The input to HTDN is text and a set of 5 images. The text goes through the Language Network $\mathcal{F}_l$ to get the language representation $h_l$ and the set of 5 images go through the Vision Network $\mathcal{F}_v$ to get the visual representation $h_v$. $h_l$ and $h_v$ are then fused together to get the multimodal representation $h_m$. The Convolutional Decision Network $\mathcal{F}_d$ conditioned on the $h_m$ makes inference about whether or not the advertisement is suspected of trafficking} 
	\label{fig:overview}
\end{figure*}

\begin{figure*}[th!]
  \centering
  % Note: raster graphics used here, because the PDF ones are extremely complicated and slow renderers to a crawl.
  \includegraphics[width=0.5\textwidth]{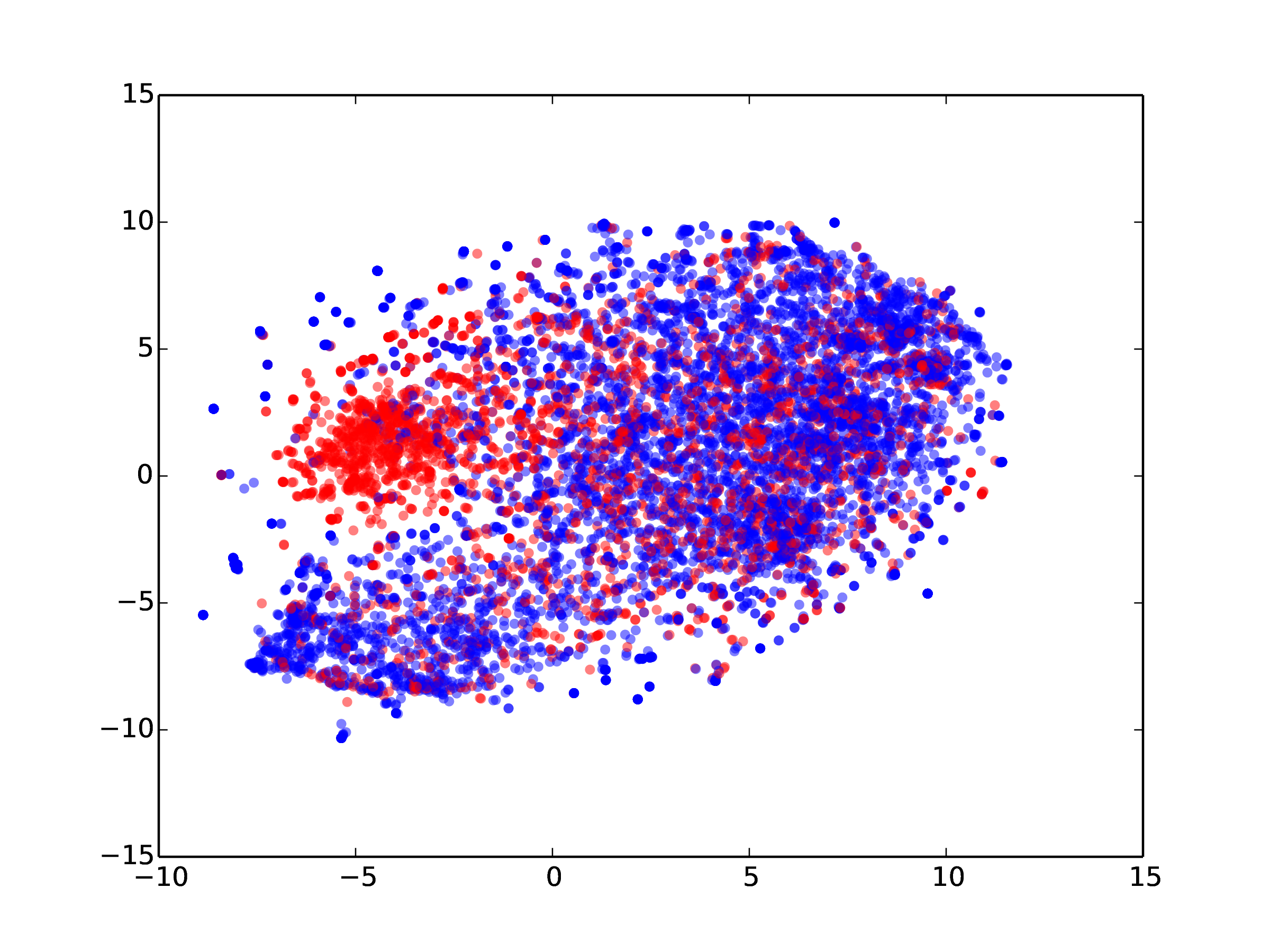}%
  \includegraphics[width=0.5\textwidth]{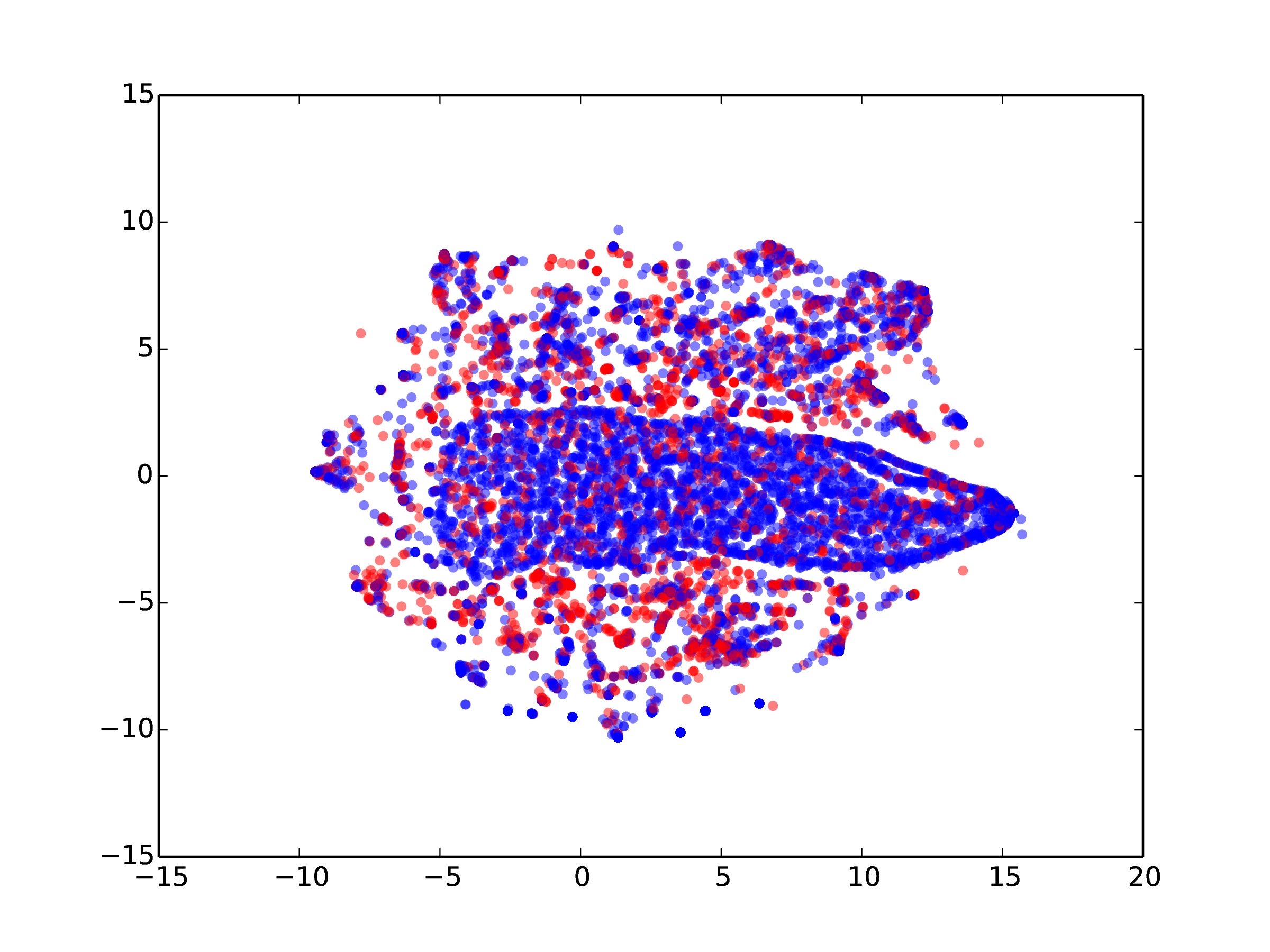}%
  \\\medskip
  \includegraphics[width=0.5\textwidth]{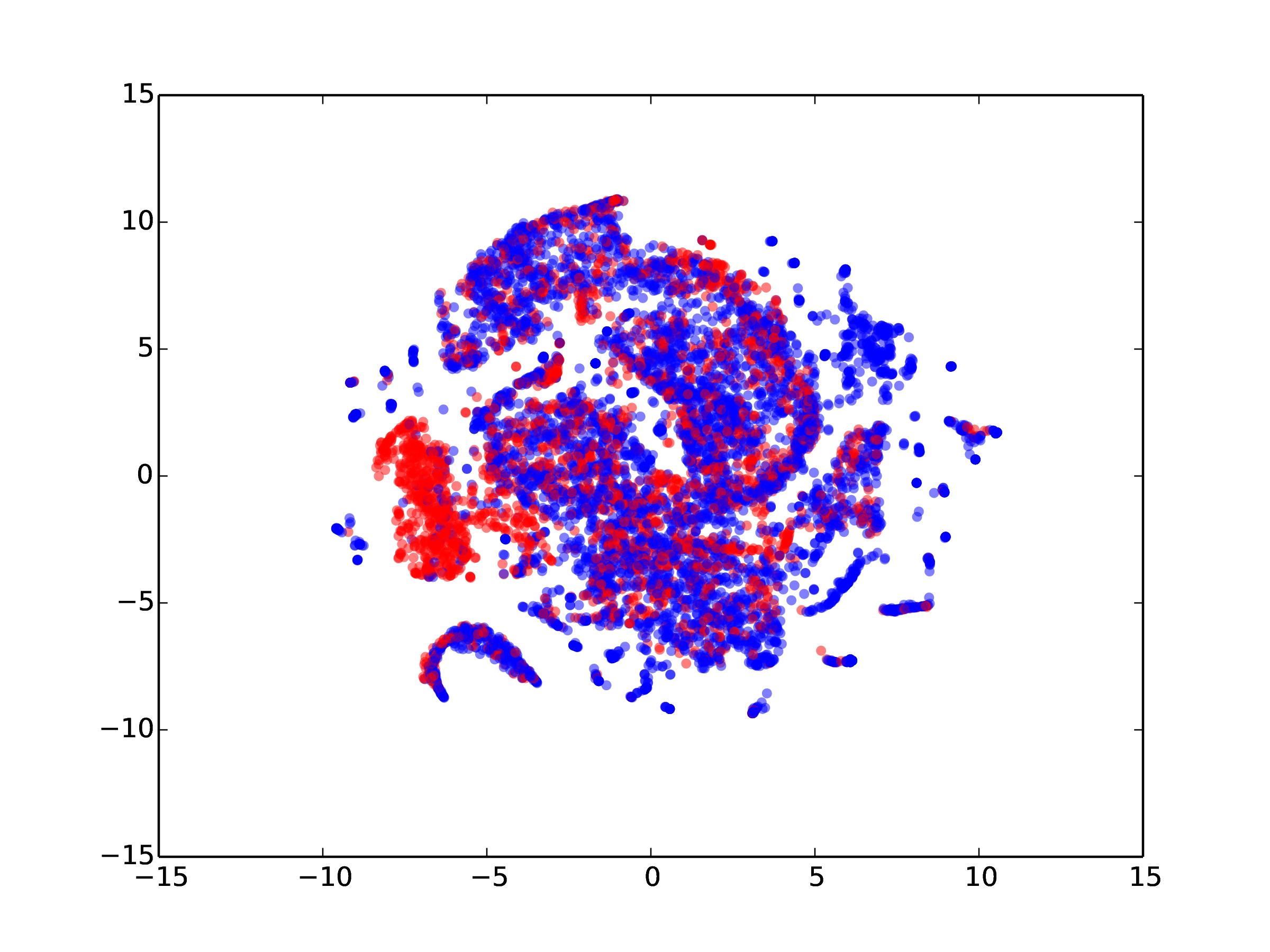}%
  \includegraphics[width=0.5\textwidth]{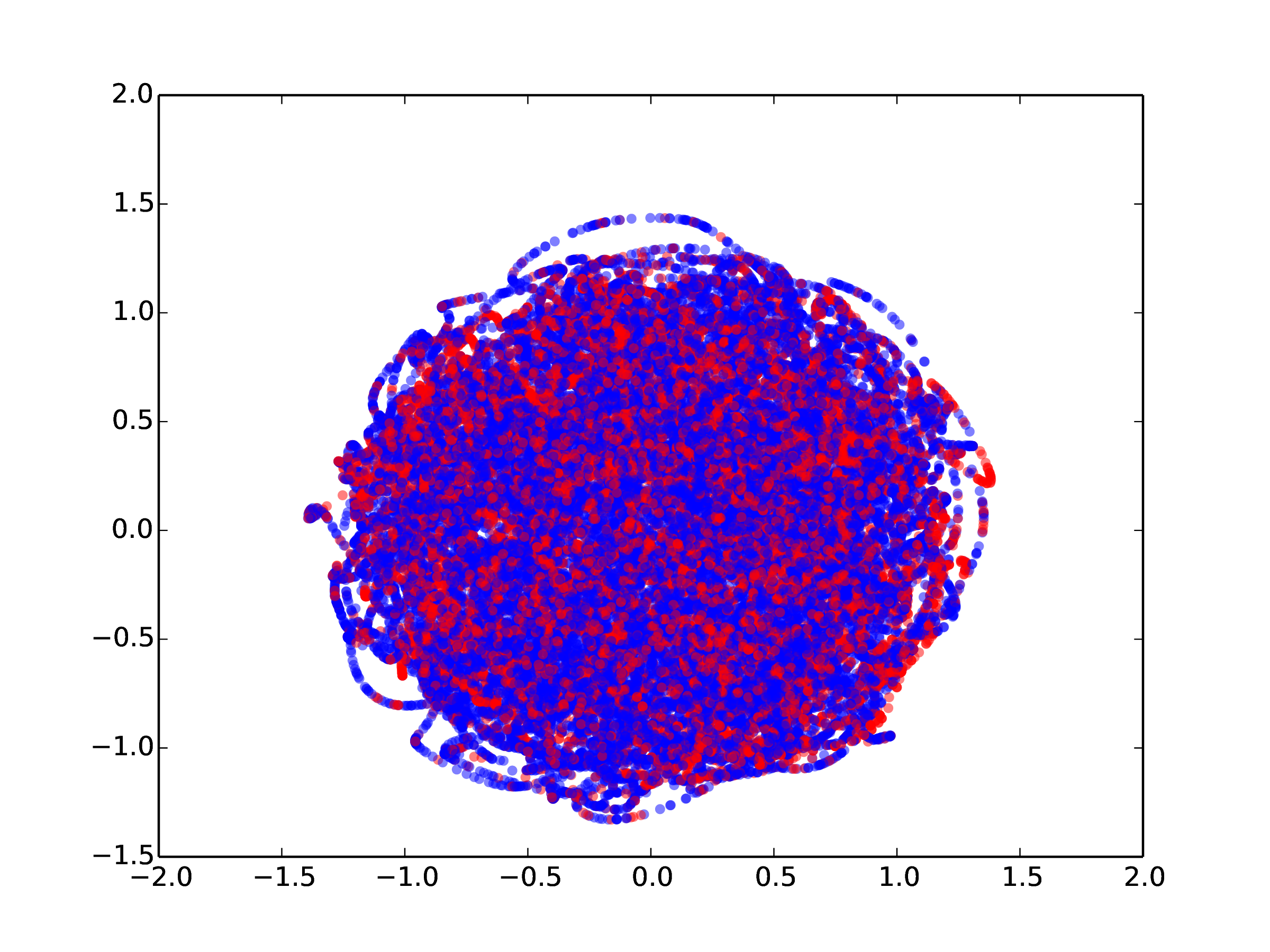}%
  \caption{%
    2D t-SNE representation of different input features for baseline models.
    Clockwise from top left: one hot vectors with expert data, one hot vectors without expert data, visual features from Vision Network $\mathcal{F}_v$, and average word vectors.
    These representations show that inference is not trivial in Trafficking-10k dataset.
  }
  \label{fig:tsne}
\end{figure*}
\section{Model}

In this section, we present our deep multimodal network called the Human Trafficking Deep Network (HTDN).
The HTDN is a multimodal network with language and vision components.
The input to the HTDN is an ad, text and images.
The HTDN is shown in Figure~\ref{fig:overview}.
In the remainder of this section, we will outline the different parts of the HTDN, and the input features to each component.

\subsection{Trafficking Word Embeddings}

Our approach to deal with the adversarial environment of escort ads is to use word vectors, defining words not based on their constituent characters, but rather based on their context.
For instance, consider the two unigrams ``cash'' and ``\copyright a\$h.''
While these contain different characters, semantically they are the same, and they occur in the same context.
Thus, our expectation is that both the unigrams will be mapped to similar vectors.
Word embeddings pre-trained on general domains do not cover most of the unigrams in Trafficking-10k.
For instance, the GloVe embedding \cite{pennington2014glove} trained on Wikipedia covers only~49.7\% of our unigrams.
The first step of the HTDN pipeline is to train word vectors \cite{mikolov2013distributed} based on the skip-gram model.
This is especially suitable for escort ads, because skip-gram models are able to capture context without relying on word order.
We train the word embedding using 1,000,000 unlabeled ads from a dataset that does not include the Trafficking-10k data.
For each advertisement, the input to the trained embedding is a sequence of words $\mathbf{\hat{w}} = [\hat{w}_1, \dotsc, \hat{w}_t]$, and the output is a sequence of $100$-dimensional word vectors $\mathbf{w} = [w_1, \dotsc, w_t]$, where $t$ is the size of the advertisement and $w_i \in \mathbb{R}^{100}$.
Our trained word vectors cover 94.9\% of the unigrams in the Trafficking-10k dataset.

\subsection{Language Network}

Our language network is designed to deal with two challenging aspects of escort advertisements: (1)~violation of constituency, and (2)~presence of irrelevant information not related to trafficking but present in ads. We address both of these issues by learning a time dependent embedding at word level. This allows the model to not rely on constituency and also remember useful information from the past, should the model get overwhelmed by irrelevant information. 
Our proposed language network, $\mathcal{F}_l$, takes as input a sequence of word vectors $\mathbf{w} = [w_1, \dotsc, w_t]$, and outputs a neural language representation $h_l$.
As a first step, $\mathcal{F}_l$ uses the word embeddings as input to a Long-Short Term Memory (LSTM) network and produces a new supervised context-aware word embedding $\mathbf{u} = [u_1, \dotsc, u_t]$ where $u_i \in \mathbb{R}^{300}$ is the output of the LSTM at time $i$.
Then, $\mathbf{u}$ is fed into a fully connected layer with dropout $p = 0.5$ to produce the neural language representation $h_l \in \mathbb{R}^{300}$ according to the following formulas with weights $W_l$ for the LSTM and implicit weights in the fully connected layers, which we represent by $\mathit{FC}$:
\begin{align}
  u_i &= \mathit{LSTM}(i, w_i; W_l) \\
  \mathbf{u} &= [u_1, \dotsc, u_t] \\
  h_l &= \mathit{FC}(\mathbf{u}).
\end{align}
The generated $h_l$ is then used as part of the HTDN pipeline, and is also trained independently to assess the performance of the language-only model.
The language network $\mathcal{F}_l$ is the combination of the LSTM and the fully-connected network.

\subsection{Vision Network}

Parallel to the language network, the vision network $\mathcal{F}_v$ takes as input advertisement images and extracts visual representations $h_v$.
The vision network takes at most five images; the median number of images per advertisement in Trafficking-10k is~$5$.
To learn contextual and abstract information from images, we use a deep convolutional neural network called Trafficking-VGG (T-VGG), a fine-tuned instance of the well-known VGG~network \cite{simonyan2014very}.
T-VGG is a deep model with $13$ consecutive convolutional layers followed by $2$ fully connected layers; it does not include the softmax layer of~VGG.
The procedure for fine-tuning T-VGG maps each individual image to a label that comes from the advertisement, and then performs end-to-end training.
For example, if there are five images in an advertisement with positive label, all five images are mapped to positive label.
After fine-tuning, three fully connected layers of 200~neurons with dropout $p = 0.5$ are added to the network.
The combination of T-VGG and the fully connected layers is the vision network $\mathcal{F}_l$.
We consider five images $\boldsymbol{\hat{\imath}} = \{ \hat{\imath}_1, \dotsc, \hat{\imath}_5 \}$ from each input advertisement.
If the advertisement has fewer than five images, zero-filled images are added.
For each image, the output of $\mathcal{F}_v$ is a representation of five images $\mathbf{i} = \{ i_1, \dotsc, i_5 \}$.
The visual representation $h_v \in \mathbb{R}^{5 \times 200}$ is a matrix with a size-$200$ representation of each of the $5$~images:
\begin{align}
  h_v = \mathcal{F}_v(\boldsymbol{\hat{\imath}}; W_v).
\end{align}

\subsection{Multimodal Fusion}

Escort advertisements have complex dynamics between text and images.
Often, neither linguistic nor visual cues alone can suffice to classify whether an ad is suspicious. Interactions between linguistic and visual cues can be non-trivial, so this requires an explicit joint representation for each neuron in the linguistic and visual representations. 
In our multimodal fusion approach we address this by calculating an outer product between language and visual representations $h_l$ and $h_v$ to build the full space of possible outcomes: 
\begin{equation}
  h_m = h_l \otimes h_v,
\end{equation}
where $\otimes$ is an outer product of the two representations.
This creates a joint multimodal tensor called $h_m$ for language and visual modalities.
In this tensor, every neuron in the language representation is multiplied by every neuron in vision representation, thus creating a new representation containing the information of both of them.
Thus, the final fusion tensor $h_m \in \mathbb{R}^{5 \times 200 \times 300}$ contains information from the joint interaction of the language and visual modalities.

\subsection{Convolutional Decision Network}

The multimodal representation $h_m$ is used as the input to the convolutional decision network $\mathcal{F}_d$.
%The first dimension in $h_m$ is considered the depth, and the other two dimensions are width and height.
$\mathcal{F}_d$ has two layers of convolution and max pooling with a dropout rate of $p = 0.5$, followed by a fully connected layer of $150$~neurons with a dropout rate of $p = 0.5$.
Performing convolutions in this space enables the model to attend to small areas of linguistic and visual cues.
It can thus find correspondences between specific combinations of the linguistic and visual representations.
The final decision is made by a single sigmoid neuron.
%with probability given by
%\begin{equation}
%  \label{eq:SIN}
%  \argmax_{\tau} \ \Pr(\tau \mid \mathbf{h}_m; W_d) = \argmax_{\tau} \ \mathcal{F}_d(\mathbf{h}_m ; W_d),
%\end{equation}
%where $\tau \in \{ 0, 1 \}$ indicates whether an advertisement is perceived as suspicious, and $W_d$ is the set of learnable weights for $\mathcal{F}_d$.
\begin{table*}[t!]
  \centering
  \newcommand{\category}[1]{\rlap{\textbf{#1}}}
  \newcommand{\entry}{\quad}
  \begin{tabular}{l*5r}
    \toprule
    \textbf{Model} & \textbf{Wt.\ Acc.\ (\%)} & \textbf{F1 (\%)} & \textbf{Acc.\ (\%)} & \textbf{Precision (\%)} & \textbf{Recall (\%)} \\
    \midrule
	\category{Random}                  & 50.0          & -          & 68.2 & - & -   \\
    \category{Keywords} \\
    \entry Random Forest              & 67.0          & 55.2          & 78.1 & 78.2 & 42.6   \\
    \entry Logistic Regression        & 69.9          & 57.8          & 78.4 & 75.5 & 46.8   \\
    \entry Linear SVM                 & 69.5          & 57.0          & 78.6 & 78.0 & 44.9   \\[1ex]
    \category{Average Trafficking Vectors} \\
    \entry Random Forest              & 67.3          & 54.1          & 78.0 & 79.3 & 41.1   \\
    \entry Logistic Regression        & 72.2          & 61.7          & 80.2 & 79.2 & 50.6   \\
    \entry Linear SVM                 & 70.3          & 57.7          & 79.2 & 80.7 & 44.9   \\[1ex]
    \category{108 One-Hot} \\
    \entry Random Forest              & 62.4          & 60.7          & 72.6 & 61.5 & 60.0   \\
    \entry Logistic Regression        & 62.5          & 45.1          & 72.2 & 60.0 & 36.1   \\
    \entry Linear SVM                 & 61.7          & 45.1          & 71.8 & 58.6 & 36.7   \\[1ex]
    \category{Bag of Words} \\
    \entry Random Forest              & 57.6          & 24.5          & 70.4 & 63.2 & 15.2   \\
    \entry Logistic Regression        & 71.1          & 24.5          & 70.4 & 63.2 & 15.2   \\
    \entry Linear SVM                 & 71.2          & 24.5          & 70.4 & 63.2 & 15.2   \\[1ex]
    \category{HTDN Unimodal} \\
    \entry $\mathcal{F}_{l}$          & 74.5          & 65.8          & 78.8 & 69.8 & 62.3   \\
    \entry $\mathcal{F}_{v}$ [VGG]    & 69.1          & 58.4          & 74.2 & 66.7 & 52.0   \\
    \entry $\mathcal{F}_{v}$ [T-VGG]  & 70.4          & {59.5}        & 77.3 & 78.3 & 48.0   \\[1ex]
    \category{HTDN}                   & \textbf{75.3} & \textbf{66.5} & 80.0 & 71.4 & 62.2   \\[1ex]
    \category{Human}                  & 83.7          & 73.7          & 84.0 & 76.7 & 70.9   \\
    \bottomrule
  \end{tabular}
  \caption{Results of our experiments. We compare our HTDN model to various baselines using different inputs. HTDN ourperforms other baselines in both weighted accuracy and F-score.}
  \label{table:results}
\end{table*}
\section{Experiments}

In our experiments, we compare the HTDN with previously used approaches for detection of trafficking suspicious ads.
Furthermore, we compare the HTDN to the performance of its unimodal components.
In all our experiments we perform binary classification of whether the advertisement is suspected of being related to trafficking.
The main comparison method that we use is the weighted accuracy and F1-score (due to imbalance it dataset).
The formulation for weighted accuracy is as follows:
\begin{equation}
  \mathrm{Wt.\ Acc.} = \frac{\mathrm{TP} \times \mathrm{N} / \mathrm{P} + \mathrm{TN}}{2 \mathrm{N}}
\end{equation}
where $\mathrm{TP}$ (resp.\ $\mathrm{TN}$) is true positive (resp.\ true negative) predictions, and $\mathrm{P}$ (resp.\ $\mathrm{N}$) is the total number of positive (resp.\ negative) examples.

\subsection{Baselines}

We compare the performance of the HTDN network with baseline models divided in 4 major categories

\textbf{Bag-of-Words Baselines.}
This set of baselines is designed to assess performance of off-the-shelf basic classifiers and basic language features.
We train random forest, logistic regression and linear SVMs to show the performance of simple language-only models.

\textbf{Keyword Baselines.}
These demonstrate the performance of models that use a set of 108 keywords, all highly related to trafficking, provided by law enforcement officers.%
\footnote{%
  Not presented in this paper due to sensitive nature of these keywords.
}
A binary one-hot vector representing these keywords is used to train the random forest, logistic regression, and linear SVM models.

\textbf{108 One-Hot Baselines.}
Similar to Keywords Baseline, we use feature selection technique to filter the most informative 108 words for detection of trafficking. We compare the performance of this baseline to Keywords baseline to evaluate the usefulness of expert knowledge in keywords selection vs automatic data-driven keyword selection.

\textbf{Average Trafficking Vectors Baselines.}
We assess the magnitude of success for the trafficking word embeddings for different classifiers.
For the random forest, logistic regression, and linear SVM models, the average word vector is calculated and used as input.

\textbf{HTDN Unimodal.}
These baselines show the performance of unimodal components of HTDN. For language we only use $\mathcal{F}_l$ component of the pipeline and for visual we use $\mathcal{F}_v$, using both pre-trained a VGG and finetuned T-VGG.

\textbf{Random and Human.}
Random is based on assigning the more frequent class in training set to all the test data, and can be considered a lower bound for our model. Human performance metrics are upper bounds for this task's metrics.

We visualize the different inputs to our baseline models to show the complexity of the dataset when using different feature sets.
Figure~\ref{fig:tsne} shows the 2D t-SNE \cite{maaten2008visualizing} representation of the training data in our dataset according to the Bag-of-Words (top right) models, expert keywords (top left), average word vectors (bottom right), and the visual representation $h_v$ bottom left.
The distribution of points suggests that none of the feature representations make the classification task trivial.

\subsection{Training Parameters}

All the models in our experiments are trained on the Trafficking-10k designated training set and tested on the designated test set.
Hyperparameter evaluation is performed using a subset of training set as validation set.
The HTDN model is trained using the Adam optimizer \cite{kingma2014adam}.
The neural weights were initialized randomly using Xavier initialization technique \cite{glorot2010understanding}.
The random forest model uses $10$~estimators, with no maximum depth, and minimum-samples-per-split value of 2.
The linear SVM model uses an $\ell_2$-penalty and a square hinge loss with $C = 1$.
\section{Results and Discussion}

The results of our experiments are shown in Table~\ref{table:results}.
We report the results on three metrics: F1-score, weighted accuracy, and accuracy.
Due to the imbalance between the numbers of positive and negative samples, weighted accuracy is more informative than unweighted accuracy, so we focus on the former.

\textbf{HTDN.}
The first observation from Table~\ref{table:results} is that the HTDN model outperforms all the proposed baselines.
There is a significant gap between the HTDN (and variants) and other non-neural approaches.
This better performance is an indicator of complex interactions in detecting dynamics of human trafficking, which is captured by the HTDN.

\textbf{Both Modalities are Helpful.}
Both modalities are helpful in predicting signs of trafficking ($\mathcal{F}_l$ and $\mathcal{F}_v$ [T-VGG]).
Fine-tuning VGG network parameters shows improvement over pre-trained VGG parameters.

\textbf{Language is More Important.}
Since $\mathcal{F}_l$ shows better performance than $\mathcal{F}_v$ [T-VGG], the language modality appears to be the more informative modality for detecting trafficking suspicious ads. 

%% This section is unsubstantiated.
%\textbf{Law Enforcement Keywords.}
%Our results indicate that reducing the feature set to set of keywords provided by the law-enforcement hurts the performance of the model.
%This indicates that keyword searching commonly used by law enforcement agencies may hurt the performance of the search queries or automatic systems currently being used by agencies.
%Also, because obfuscation is widely used in escort advertisement websites, these keywords may not generalize.
%The same is true for feature set reduction using feature selection. 
\section{Conclusion and Future Work}

In this paper, we took a major step in multimodal modeling of suspected online trafficking advertisements.
We presented a novel dataset, Trafficking-10k, with more than 10,000 advertisements annotated for this task.
The dataset contains two modalities of information per advertisement: text and images.
We designed a deep multimodal model called the Human Trafficking Deep Network (HTDN).
We compared the performance of the HTDN to various models that use language and vision alone.
The HTDN outperformed all of these, indicating that using information from both sources may be more helpful than using just one.

\textbf{Exploring language through character modeling.}
In order to eliminate the need for retraining the word vectors as the language of the domain evolves, we plan to use character models to learn a better language model for trafficking.
As new obfuscated words are introduced in escort advertisements, our hope is that character models will stay invariant to these obfuscations.
%One way to do so is learning the parameters of the language network with presence of manual obfuscation in the training data.
% ^ What does this mean? :(

\textbf{Understanding images.}
While CNNs have proven to be useful for many different computer vision tasks, we seek to improve the learning capability of the visual network.
Future direction involves using graphical modeling to understand interactions in the scene.
Another direction involves working to understand text in images, which can provide more information about the subjects of the images.

Given that the current state of the art in this area generally does not use deep models, this may be a major opportunity for improvement.
To this end, we encourage the research community to reach out to Cara Jones, an author of this paper, to obtain a copy of Trafficking-10k and other training data.
\section*{Acknowledgements}

We would like to thank William Chargin for creating figures and revising this paper.
We would also like to thank Torsten W\"ortwein for his assistance in visualizing our data.
Furthermore, we would like to thank our anonymous reviewers for their valuable feedback.
Finally, we would like to acknowledge collaborators from Marinus Analytics for the time and effort that they put into annotating advertisements for the dataset, and for allowing us to use their advertisement data.

\bibliography{acl2017}

\begin{thebibliography}{}
\expandafter\ifx\csname natexlab\endcsname\relax\def\natexlab#1{#1}\fi

\bibitem[{Antol et~al.(2015)Antol, Agrawal, Lu, Mitchell, Batra,
  Lawrence~Zitnick, and Parikh}]{antol2015vqa}
Stanislaw Antol, Aishwarya Agrawal, Jiasen Lu, Margaret Mitchell, Dhruv Batra,
  C~Lawrence~Zitnick, and Devi Parikh. 2015.
\newblock Vqa: Visual question answering.
\newblock In {\em Proceedings of the IEEE International Conference on Computer
  Vision\/}. pages 2425--2433.

\bibitem[{Donahue et~al.(2015)Donahue, Anne~Hendricks, Guadarrama, Rohrbach,
  Venugopalan, Saenko, and Darrell}]{donahue2015long}
Jeffrey Donahue, Lisa Anne~Hendricks, Sergio Guadarrama, Marcus Rohrbach,
  Subhashini Venugopalan, Kate Saenko, and Trevor Darrell. 2015.
\newblock Long-term recurrent convolutional networks for visual recognition and
  description.
\newblock In {\em Proceedings of the IEEE conference on computer vision and
  pattern recognition\/}. pages 2625--2634.

\bibitem[{Dubrawski et~al.(2015)Dubrawski, Miller, Barnes, Boecking, and
  Kennedy}]{dubrawski2015leveraging}
Artur Dubrawski, Kyle Miller, Matthew Barnes, Benedikt Boecking, and Emily
  Kennedy. 2015.
\newblock Leveraging publicly available data to discern patterns of
  human-trafficking activity.
\newblock {\em Journal of Human Trafficking\/} 1(1):65--85.

\bibitem[{Glorot and Bengio(2010)}]{glorot2010understanding}
Xavier Glorot and Yoshua Bengio. 2010.
\newblock Understanding the difficulty of training deep feedforward neural
  networks.
\newblock In {\em Aistats\/}. volume~9, pages 249--256.

\bibitem[{Hotline(2017)}]{hotline2017}
National Human~Trafficking Hotline. 2017.
\newblock \href{https://humantraffickinghotline.org/states}{Hotline
  statistics}.
\newblock
  \href{https://humantraffickinghotline.org/states}{https://humantraffickinghotline.org/states}.

\bibitem[{Kennedy(2012)}]{kennedy2012predictive}
Emily Kennedy. 2012.
\newblock Predictive patterns of sex trafficking online.
\newblock {\em Dietrich College Honors Theses\/} .

\bibitem[{Kingma and Ba(2014)}]{kingma2014adam}
Diederik Kingma and Jimmy Ba. 2014.
\newblock Adam: A method for stochastic optimization.
\newblock {\em arXiv preprint arXiv:1412.6980\/} .

\bibitem[{Maaten and Hinton(2008)}]{maaten2008visualizing}
Laurens van~der Maaten and Geoffrey Hinton. 2008.
\newblock Visualizing data using t-sne.
\newblock {\em Journal of Machine Learning Research\/} 9(Nov):2579--2605.

\bibitem[{McCarthy(2014)}]{mccarthy2014human}
Lauren~A McCarthy. 2014.
\newblock Human trafficking and the new slavery.
\newblock {\em Annual Review of Law and Social Science\/} 10:221--242.

\bibitem[{Mikolov et~al.(2013)Mikolov, Sutskever, Chen, Corrado, and
  Dean}]{mikolov2013distributed}
Tomas Mikolov, Ilya Sutskever, Kai Chen, Greg~S Corrado, and Jeff Dean. 2013.
\newblock Distributed representations of words and phrases and their
  compositionality.
\newblock In {\em Advances in neural information processing systems\/}. pages
  3111--3119.

\bibitem[{Nagpal et~al.(2015)Nagpal, Miller, Boecking, and
  Dubrawski}]{nagpal2015entity}
Chirag Nagpal, Kyle Miller, Benedikt Boecking, and Artur Dubrawski. 2015.
\newblock An entity resolution approach to isolate instances of human
  trafficking online.
\newblock {\em arXiv preprint arXiv:1509.06659\/} .

\bibitem[{Pennington et~al.(2014)Pennington, Socher, and
  Manning}]{pennington2014glove}
Jeffrey Pennington, Richard Socher, and Christopher~D Manning. 2014.
\newblock {GloVe}: Global vectors for word representation.
\newblock In {\em EMNLP\/}. volume~14, pages 1532--1543.

\bibitem[{Poria et~al.(2017)Poria, Cambria, Bajpai, and
  Hussain}]{poria2017review}
Soujanya Poria, Erik Cambria, Rajiv Bajpai, and Amir Hussain. 2017.
\newblock A review of affective computing: From unimodal analysis to multimodal
  fusion.
\newblock {\em Information Fusion\/} 1:34.

\bibitem[{Poria et~al.(2016)Poria, Chaturvedi, Cambria, and
  Hussain}]{poria2016convolutional}
Soujanya Poria, Iti Chaturvedi, Erik Cambria, and Amir Hussain. 2016.
\newblock Convolutional mkl based multimodal emotion recognition and sentiment
  analysis.
\newblock In {\em 2016 IEEE 16th International Conference on Data Mining
  (ICDM)\/}. IEEE, pages 439--448.

\bibitem[{Scherer et~al.(2016)Scherer, Lucas, Gratch, Rizzo, and
  Morency}]{scherer2016self}
Stefan Scherer, Gale~M Lucas, Jonathan Gratch, Albert~Skip Rizzo, and
  Louis-Philippe Morency. 2016.
\newblock Self-reported symptoms of depression and ptsd are associated with
  reduced vowel space in screening interviews.
\newblock {\em IEEE Transactions on Affective Computing\/} 7(1):59--73.

\bibitem[{Simonyan and Zisserman(2014)}]{simonyan2014very}
Karen Simonyan and Andrew Zisserman. 2014.
\newblock Very deep convolutional networks for large-scale image recognition.
\newblock {\em arXiv preprint arXiv:1409.1556\/} .

\bibitem[{Specia et~al.(2016)Specia, Frank, Sima'an, and
  Elliott}]{specia2016shared}
Lucia Specia, Stella Frank, Khalil Sima'an, and Desmond Elliott. 2016.
\newblock A shared task on multimodal machine translation and crosslingual
  image description.
\newblock In {\em Proceedings of the First Conference on Machine Translation,
  Berlin, Germany. Association for Computational Linguistics\/}.

\bibitem[{UNODC(2008)}]{unodc2008}
UNODC. 2008.
\newblock
  \href{http://www.ungift.org/doc/knowledgehub/resource-centre/GIFT_Human_Trafficking_An_Overview_2008.pdf}{Human
  trafficking: An overview}.
\newblock Web, New York.
\newblock
  \href{http://www.ungift.org/doc/knowledgehub/resource-centre/GIFT_Human_Trafficking_An_Overview_2008.pdf}{http://www.ungift.org/doc/knowledgehub/resource-centre/GIFT_Human_Trafficking_An_Overview_2008.pdf}.

\bibitem[{Valstar et~al.(2016)Valstar, Gratch, Schuller, Ringeval, Lalanne,
  Torres~Torres, Scherer, Stratou, Cowie, and Pantic}]{valstar2016avec}
Michel Valstar, Jonathan Gratch, Bj{\"o}rn Schuller, Fabien Ringeval, Dennis
  Lalanne, Mercedes Torres~Torres, Stefan Scherer, Giota Stratou, Roddy Cowie,
  and Maja Pantic. 2016.
\newblock Avec 2016: Depression, mood, and emotion recognition workshop and
  challenge.
\newblock In {\em Proceedings of the 6th International Workshop on Audio/Visual
  Emotion Challenge\/}. ACM, pages 3--10.

\bibitem[{Venek et~al.(2016)Venek, Scherer, Morency, Rizzo, and
  Pestian}]{venek2016adolescent}
Verena Venek, Stefan Scherer, Louis-Philippe Morency, Albert Rizzo, and John
  Pestian. 2016.
\newblock Adolescent suicidal risk assessment in clinician-patient interaction.
\newblock {\em IEEE Transactions on Affective Computing\/} .

\bibitem[{You et~al.(2016)You, Jin, Wang, Fang, and Luo}]{you2016image}
Quanzeng You, Hailin Jin, Zhaowen Wang, Chen Fang, and Jiebo Luo. 2016.
\newblock Image captioning with semantic attention.
\newblock In {\em Proceedings of the IEEE Conference on Computer Vision and
  Pattern Recognition\/}. pages 4651--4659.

\bibitem[{Yu et~al.(2013)Yu, Scherer, Devault, Gratch, Stratou, Morency, and
  Cassell}]{yu2013multimodal}
Zhou Yu, Stefen Scherer, David Devault, Jonathan Gratch, Giota Stratou,
  Louis-Philippe Morency, and Justine Cassell. 2013.
\newblock Multimodal prediction of psychological disorders: Learning verbal and
  nonverbal commonalities in adjacency pairs.
\newblock In {\em Semdial 2013 DialDam: Proceedings of the 17th Workshop on the
  Semantics and Pragmatics of Dialogue\/}. pages 160--169.

\bibitem[{Zadeh et~al.(2016{\natexlab{a}})Zadeh, Zellers, Pincus, and
  Morency}]{zadeh2016mosi}
Amir Zadeh, Rowan Zellers, Eli Pincus, and Louis-Philippe Morency.
  2016{\natexlab{a}}.
\newblock Mosi: Multimodal corpus of sentiment intensity and subjectivity
  analysis in online opinion videos.
\newblock {\em arXiv preprint arXiv:1606.06259\/} .

\bibitem[{Zadeh et~al.(2016{\natexlab{b}})Zadeh, Zellers, Pincus, and
  Morency}]{zadeh2016multimodal}
Amir Zadeh, Rowan Zellers, Eli Pincus, and Louis-Philippe Morency.
  2016{\natexlab{b}}.
\newblock Multimodal sentiment intensity analysis in videos: Facial gestures
  and verbal messages.
\newblock {\em IEEE Intelligent Systems\/} 31(6):82--88.

\bibitem[{Zhou et~al.(2016)Zhou, Luo, and McGibbney}]{zhou2016multimedia}
Andrew~Jie Zhou, Jiyun Luo, and Lewis~John McGibbney. 2016.
\newblock Multimedia metadata-based forensics in human trafficking web data.
\newblock {\em Vanessa Murdock, Charles LA Clarke, Jaap\/} page~10.

\end{thebibliography}
\bibliographystyle{acl_natbib}

\end{document}